\documentclass[10pt,letterpaper]{article}

\usepackage{cogsci}

 \cogscifinalcopy 

\usepackage{float}

\author{
Sasha Boguraev$^{*}$ \quad Qing Yao$^{*}$ \quad Kyle Mahowald\\
Department of Linguistics, The University of Texas at Austin\\
\texttt{\{sasha.boguraev,qyao,kyle\}@utexas.edu}\\
$^{*}$These authors contributed equally.
}

\usepackage{tabularx,ragged2e,booktabs,caption}
\usepackage{subcaption}
\usepackage{graphicx}
\usepackage{linguex}
\usepackage{natbib}
\usepackage[table, dvipsnames]{xcolor}
\usepackage{arydshln}
\usepackage{tikz-qtree}
\usepackage{amsmath}
\usepackage{bbding}
\usepackage{pifont}
\usepackage{booktabs}
\usepackage{csquotes}
\usepackage{comment}
\usepackage{pgfplots}
\pgfplotsset{compat=1.16}
\usepackage{epigraph} 
\usepackage{tikz-dependency}

\usepackage[textsize=scriptsize,textwidth=2cm]{todonotes}

\usepackage{listings,xcolor}
\usepackage{tcolorbox}

\usepackage{enumitem}
\usepackage{wrapfig}

\usetikzlibrary{patterns,arrows.meta,decorations.pathmorphing}
\definecolor{lightgreen}{RGB}{200,230,200}
\definecolor{midgreen}{RGB}{150,200,150}
\definecolor{darkgreen}{RGB}{100,170,100}
\definecolor{lightblue}{RGB}{190,200,230}

\definecolor{codegray}{rgb}{0.95,0.95,0.95}
\lstnewenvironment{longprompt}[1][]
{\lstset{basicstyle={\small\it},prebreak={},postbreak=\mbox{\hspace{-2.25 em}},backgroundcolor=\color{codegray},frame=single,framerule=1pt,#1}}
{}

\usepackage{nameref}

\title{France or Spain or Germany or France:\\A Neural Account of Non-Redundant Redundant Disjunctions
}

\begin{document}
\setlength{\Exlabelwidth}{1em}
\setlength{\Exlabelsep}{1em}
\setlength{\SubExleftmargin}{1.5em}

\maketitle

\begin{abstract}
Sentences like ``She will go to France or Spain, or perhaps to Germany or France.'' appear formally redundant, yet become acceptable in contexts such as ``Mary will go to a philosophy program in France or Spain, or a mathematics program in Germany or France.'' While this phenomenon has typically been analyzed using symbolic formal representations, we aim to provide an account grounded in artificial neural mechanisms.
We first present new behavioral evidence from humans and large language models demonstrating the robustness of this apparent non-redundancy across contexts. We then show that, in language models, redundancy avoidance arises from two interacting mechanisms: models learn to bind contextually relevant information to repeated lexical items, and Transformer induction heads selectively attend to these context-licensed representations. We argue that this neural explanation sheds light on the mechanisms underlying context-sensitive semantic interpretation, and that it complements existing symbolic analyses.

\end{abstract}

\noindent 

\section{Introduction}
\label{sec:introduction}

If ``or'' has its standard literal logical meaning, then it seems redundant to say:

\ex. She will go to France or Spain, or perhaps to Germany or France.\label{ex:bare}

It's redundant because it says ``France'' twice, and thus where she will go is logically equivalent to the transparently redundant $A \vee B \vee C \vee A$. 
But consider that same sentence with additional context: 

\ex. Mary will go to a philosophy program in France or Spain, or a mathematics program in Germany or France. She will go to France or Spain, or perhaps to Germany or France.\label{ex:critical}

In the latter case, it doesn't seem redundant even though ``France'' is repeated.\footnote{The particular test case we focus on is from an earlier version of \citet{mandelkern2024disjunction}, since it is most amenable to our computational analysis. Mandelkern credits Nathan Klinedinst for some of the empirical observations that we draw on here.}

\begin{figure}
    \centering
    \includegraphics[width=\linewidth]{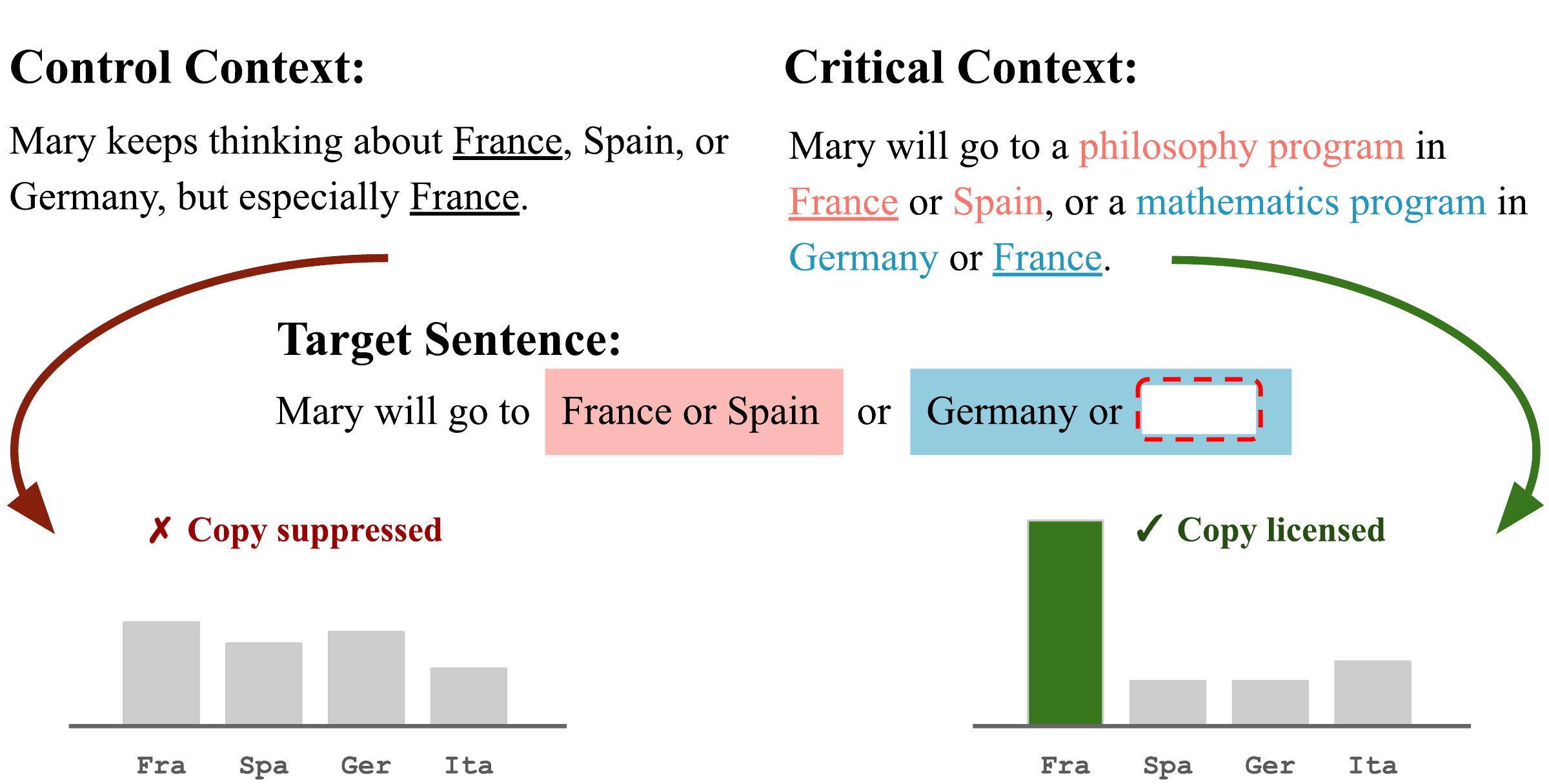}
    \caption{In context, our critical sentences become non-redundant and large LMs reliably produce the repeated item; in the control condition, they suppress the copy mechanism.}
    \label{fig:fig1}
    \vspace{-10pt}
\end{figure}

Thus, a language user (whether human or LM) has to learn that it's natural to repeat ``France'' after the appropriate context.
\citet{mandelkern2024disjunction} gives an elegant formal analysis of these phenomena, arguing for handling them semantically rather than pragmatically \citep[building on work by][]{zimmermann2000free,geurts2005entertaining,goldstein2019free}.
In particular, he argues that the possibility component associated with disjunction must be part of its semantic (as opposed to pragmatic) contribution.
On his proposal, a sentence of the form $\textit{p or q}$ is interpreted as $(p \vee q) \wedge \Diamond p \wedge \Diamond q$, where the $\Diamond$ is a possibility modal that can be epistemic, circumstantial, or deontic.

Our position, as articulated in more depth in recent work on Language Models and linguistics \citep{futrell2025linguistics}, is that it should be possible to give an account of linguistic phenomena (including this one) in the language of neural interpretability, just as it is possible to do so in the language of formal semantics.
After presenting an LM-based account of syntactic phenomena, the linguist Chris Potts said: ``It seems to me that I can talk about the feature in either way, symbolic or neural, without any loss of accuracy'' \citep{potts2025}.
That is our goal: a neural account of the phenomenon that formal semantics has treated symbolically, one that explains the same behavioral pattern through the representational resources of neural networks, rather than traditional formalism.

To be clear, our goal is not to claim that such a neural account validates or invalidates any particular formal semantic analysis, but to give an alternative representation in the language of neural interpretability, as opposed to in the language of formal semantics.

The neural account of the phenomena that we hypothesize is as follows: First, in an example like \ref{ex:critical}, the first sentence causes the nouns ``Spain'' and the first ``France''  to become contextually bound to the ``philosophy program'' and ``Germany'' and the second ``France'' to the ``mathematics program''. Thus, there are now two polysemous ``Frances'': both of which retain some of their original meaning in neural space but which are now also imbued with the contextually relevant information. 
In LM terms, we can think of this as a kind of in-context learning, consistent with a large literature showing that context can significantly shift the representation of particular tokens \citep{brown2020language,olsson2022incontextlearninginductionheads}.

Second, drawing on work in mechanistic interpretability, we hypothesize that the correct completion of sentences like \ref{ex:critical} will involve sensitively deploying induction heads to copy the relevant tokens from the first sentence \citep{elhage2021mathematical, olsson2022incontextlearninginductionheads, hendel2023context, todd2024function,feucht2025dualroute}.
Induction heads, which allow the copying of information in neural models, are perhaps the best understood aspect of neural network interpretability.
Thus, taken together, these two factors (the desired behavior's sensitivity to preceding context in the prompt, and the fact that the phenomenon involves either the activation or suppression of repetition via induction heads) make the phenomenon a particularly promising test case for these methods.

We offer this account as its own linguistic analysis, one that is non-traditional but that serves many of the goals of traditional linguistic analysis insofar as it gives an explanatory and mechanistic account of a particular linguistic phenomenon.
We remain agnostic as to whether it is underlyingly isomorphic in some way to formal semantic accounts of this phenomenon.
But because LMs handle the phenomenon empirically (as we show) and we can analyze their behavior mechanistically, we take this approach to be promising.

To that end, in this paper we first present human behavioral judgments showing that people are sensitive to these phenomena, and that sufficiently large LMs reproduce these patterns while smaller ones struggle.
We then turn to mechanistic analyses: activation patching reveals that the key disjuncts become contextually laden in ways that causally affect copying behavior, and analysis of induction heads \citep{feucht2025dualroute} shows how these heads are selectively deployed or suppressed.

\section{Experiment 1: Behavioral Performance in Humans and Models}

First, we provide novel empirical support for the phenomenon of non-redundant redundant disjunctions in humans, before replicating it in LMs.

\paragraph{Materials.}

Each item involves three entities X, Y, and Z drawn from 9 semantic domains (countries, cities, jobs, courses, instruments, sports, languages, drinks, and hobbies), a named agent drawn from 32 English names, and a domain-appropriate verb with two contextual suffixes that provide distinct possibility conditions (e.g., \textit{relocate to \ldots\ temporarily/permanently} for countries; \textit{play \ldots\ for fun/professionally} for instruments).

\paragraph{Critical items.}\label{sec:critical-items}
Critical items follow the pattern in \ref{ex:critical}: Sentence 1 (S1) establishes a structured disjunctive context where entity X appears in both disjunctions, bound to different contextual modifiers. Sentence 2 (S2) is then presented as a completion task, where the expected completion is X---the repeated item which is non-redundant given the context in S1.

We manipulated three binary factors in a $2 \times 2 \times 2$ factorial design, all varying the structure of S1 while holding S2 fixed:
\begin{itemize}[nosep]
    \item \textbf{First disjunction order}: whether S1's first disjunction is \textit{X or Y} (matching S2) or \textit{Y or X}.
    \item \textbf{Second disjunction order}: whether S1's second disjunction is \textit{Z or X} (matching) or \textit{X or Z}.
    \item \textbf{Halves order}: whether the two halves of S1 appear in the same order as the corresponding portions of S2, or are swapped.
\end{itemize}

\paragraph{Control items.}
Control items replace S1 with a simple list statement mentioning X, Y, and Z without disjunctive structure, but in which one of the nouns appears twice:

\ex. \textit{Lucas keeps thinking about visiting France, Germany, and Italy, but especially France.}\label{ex:control}

\noindent S2 is identical to the critical conditions. 
In the absence of structured disjunctions in S1, repeating X in S2 is genuinely redundant.
If the effect arises merely from repeating the noun, we would expect to see repetition in the control.
Note that the control removes both the disjunctive structure and the distinct contextual modifiers simultaneously---this is by design, since on Mandelkern's analysis it is the disjunctive structure that creates distinct possibility conditions in the first place. 
The two are not independent.\footnote{A reviewer asked why the control is not of the form: ``Lucas keeps thinking about visiting France or Germany, or Italy or France.  A friend asks Lucas which places are possibilities.  Lucas replies `I will visit France or Germany, or Italy or \_\_\_.''' where there is identical disjunctive context in S1 and S2, but without context. We see this, though, as implicitly creating distinct context between the first pair and second pair, and so predict it would behave as a critical item and not a control.} 

\subsection{Experiment 1a: Comparing Human Performance on Critical vs. Control Stimuli}

\paragraph{Participants.}
We recruited 243 participants via Prolific.
We excluded 6 participants who failed at least one attention check and 1 participant with missing data, leaving 236 for analysis.

\paragraph{Materials.}
We randomly sampled a dataset of critical and control items according to the above procedure, but with additional ``bridging'' context in critical items to make S2 a more natural continuation of S1 (an example shown in \ref{ex:with-bridge}).
Each participant saw one critical trial and one control trial. 

Critical items were evenly distributed across the 8 factorial conditions in a Latin square design, yielding approximately 15 participants per condition.
Presentation order was counterbalanced: roughly half the participants saw the control item first and the critical item second, and vice versa.

\ex. \textit{Lucas will relocate to France or Germany early in life or Italy or France later in life. A friend asks Lucas which places are possibilities. Lucas replies: ``I will relocate to France or Germany or Italy or \_\_\_\_}\label{ex:with-bridge}

In addition to the experimental items, each participant completed 4 filler items and 3 attention checks consisting of simple sentence completions.

\paragraph{Procedure.}
The task was presented as a free-response sentence completion.
Participants read S1, additional context to make S2 more natural, then S2, and were tasked with typing a completion of S2.
Responses were deemed ``correct'' in the critical condition if the participant produced the repeated item X, and in the control condition if they produced a word \textit{not} among \{X, Y, Z\}. Note that ``correct'' here refers to the behavior of interest, not  prescriptive notion of correctness.

\paragraph{Analysis.}
We restrict our analysis to each participant's first trial, due to strong and undesirable priming effects from the order in which control and critical stimuli were seen, yielding 119 critical-first participants and 117 control-first participants.

\paragraph{Results.}
Participants produced the repeated item X at a much higher rate in the critical condition than in the control condition: 72.3\% (86/119) vs.\ 28.2\% (33/117; by proportion test, $\chi^2 = 45.82$, $p < .001$).
That is, when S1 contained structured disjunctions binding X to distinct possibility conditions, participants overwhelmingly completed S2 with the repeated item, whereas when S1 merely listed X, Y, and Z without disjunctive structure, participants tended to produce a novel item instead. In sum, humans robustly show the predicted pattern: structured disjunctive context à la \citet{mandelkern2024disjunction} licenses repetition, while a neutral framing does not.

\begin{figure*}[t]
    \centering
    \begin{subfigure}[b]{0.48\linewidth}
        \centering
        \includegraphics[width=\linewidth]{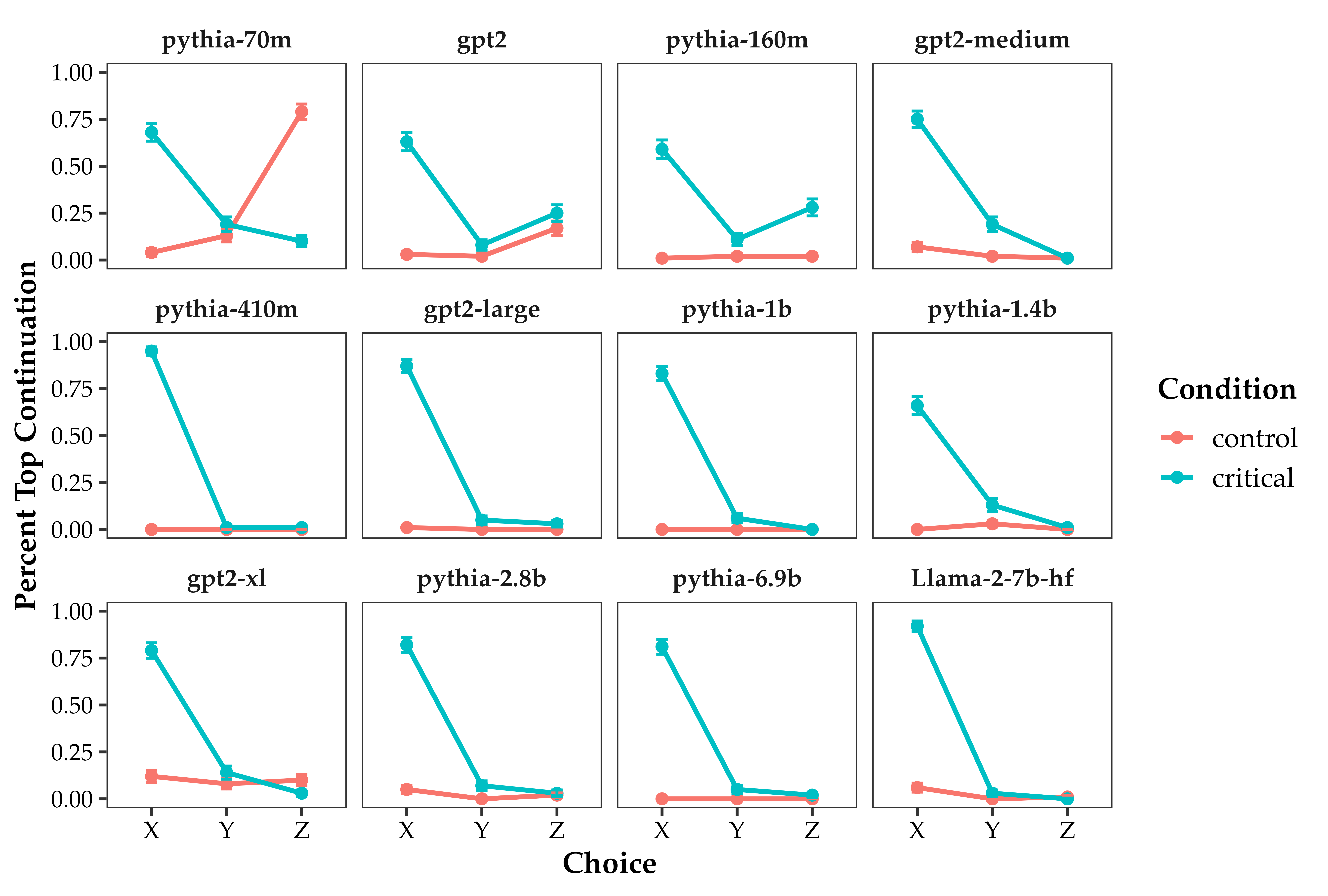}
        \caption{Generation rate for each of the three items in a given disjunction, with facets ordered by model size.}
        \label{fig:by-model}
    \end{subfigure}
    \hfill
    \begin{subfigure}[b]{0.48\linewidth}
        \centering
        \includegraphics[width=\linewidth]{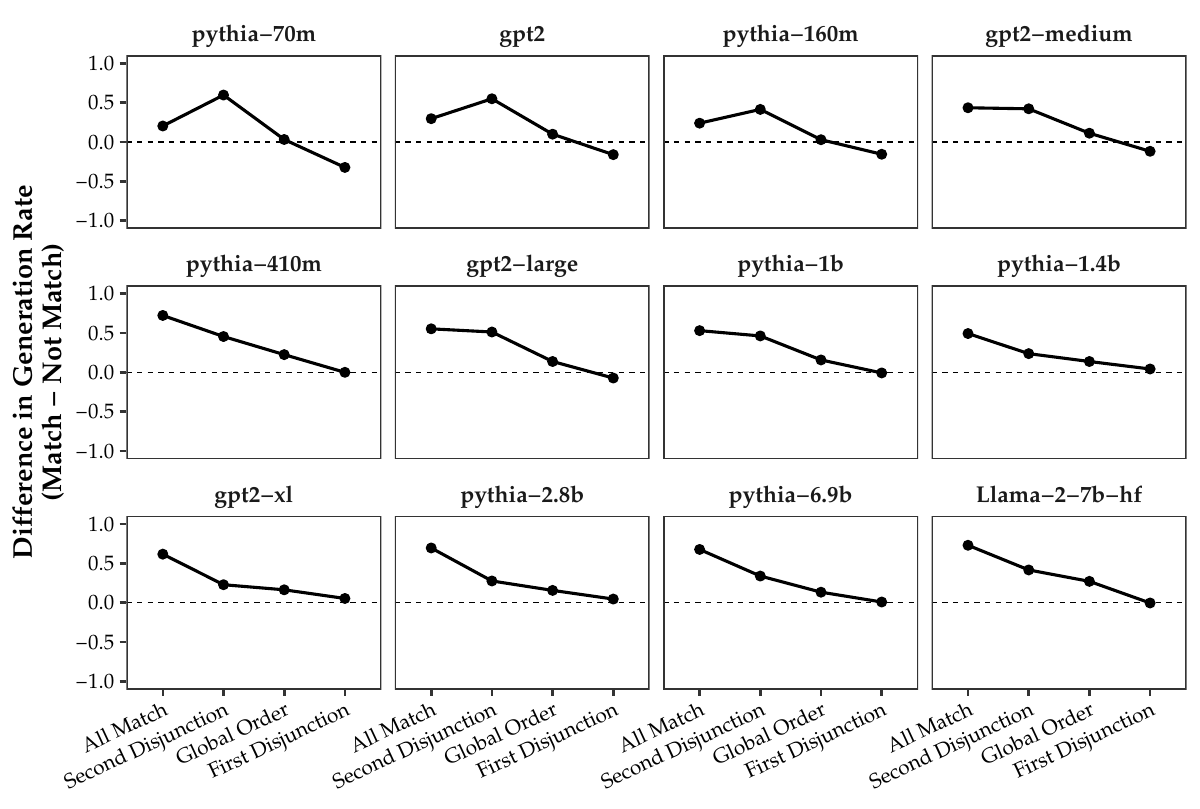}
        \caption{Difference in generation rate between critical and controls over experimental conditions.}
        \label{fig:by-model-ordering}
    \end{subfigure}
    \caption{\textbf{Left: } We broadly see LM competency, with high rate of the repeated answer (X) as top choice. \textbf{Right: } We also see order preferences in LMs. Namely in small models we see a strong preference towards a match in the ordering of the second disjunction, but as models get larger this preference gets superseded by a preference for total matching between S1 and S2.}
    \label{fig:side-by-side}
\end{figure*}

\subsection{Experiment 1b: LM Behavior}

Having established the main effect in humans, we replicated it for LMs.

\paragraph{Models.} We use 12 autoregressive language models across three model families and a range of scales:
\begin{itemize}[nosep]
    \item \textbf{Pythia} \citep{biderman}: 70M, 160M, 410M, 1B, 1.4B, 2.8B, 6.9B
    \item \textbf{GPT-2} \citep{radford2019language}: Small (124M), Medium (355M), Large (774M), XL (1.5B)
    \item \textbf{Llama} \citep{touvron2023llama2openfoundation}: Llama2-7B
\end{itemize}

\paragraph{Procedure.}\label{sec:procedure}
For experiments with LMs, we sampled 100 critical items in each of the 8 shuffled conditions and 100 control items that share the same domain and entities. For each item, the entities X, Y, and Z are single-tokens to maximize induction effects.
For each trial, we concatenated S1 and the incomplete S2, then measured the percent of items for which each of X, Y, or Z is the most probable next token for a given model (which we term ``generation rate''). This is an argmax-based metric: for each item, we check which of the three candidates has the highest probability and then report the proportion of items where each candidate wins.

\paragraph{Results.} Our results can be seen in Figure \ref{fig:by-model}. LMs show competency on this task, with models generally producing the repeated entity in the critical case, and no repetitions in the control case. Furthermore, we see moderate scaling effects, where the smallest models (under 400M parameters) perform worse than those above this threshold (with the exception of \texttt{pythia-1.4b}).
The \texttt{pythia-70m} model, for instance, shows apparent ``dumb copying'' behavior in the controls, repeating the most recent noun (Z), e.g., ``She will go to France or Spain or Germany or Germany''.  

We also analyze the effect of ordering in LMs. Figure \ref{fig:by-model-ordering} displays the difference in generation rate of the correct answer (X) between the critical sentence and control sentence in four groups. These conditions represent whether a match occurs in the first disjunction, the second disjunction, the global ordering of disjunctions, or a full match across all three.
Across our LMs, the order of the second disjunction is the most influential of our experimental factors with small models displaying the largest difference in generation rate. However, as we move to large models, a match in all orderings has a larger difference than any of the factors individually. This tradeoff happens at a similar boundary (400M parameters) as we see in our previous result, suggesting that the scaling trends and this phenomenon may be related.

We further validate these results by fitting a mixed effect logistic regression.
We predicted a binary indicator variable for ``correctly'' generating X, with fixed effect terms for (a) whether the order fully matched in S1 and S2 (e.g., XYZX) or was in some way permuted, (b) the model size (in billions of parameters), and (c) their interaction. 
 We included random effects by item \citep[with maximal random effects as per][]{barr_random_2013}.
We found significant negative effects for a mismatch ($\beta = -1.88$, $p<.001$), significant positive effects for model size (in billions of parameters; $\beta = 0.21$, $p<.01$), and significant negative effects for their interaction ($\beta = -0.47$, $p<.001$).
This suggests changing the order decreases critical repetitions.
Inspection of Figure \ref{fig:by-model-ordering} further suggests that disruption to ordering is particularly great when the second disjunction is permuted, followed by the global order, and then the first disjunction mattering least.

\subsection{Experiment 1c: Human Ordering Effects}

Given the strong ordering effects in LMs, we investigated whether humans show analogous sensitivity. We hypothesize that disjunctions may be influenced by well-established ordering effects in linguistics, particularly of the kind emphasized in the construction grammar and usage-based traditions \citep{goldberg1995constructions,goldberg2006constructions}.
Even seemingly logically equivalent forms like ``A or B'' vs. ``B or A'' can be different constructions. For instance, ``odds and ends'' conveys something different than ``ends and odds'' \citep{malkiel1959studies}.
Plausibly, when in \ref{ex:critical}, ``Germany or France'' is treated as an on-the-fly construction associated with the ``mathematics program'', whereas the completion ``France or Germany'' would not be as strongly associated with mathematics.

To home in on ordering effects, we conducted another study via Prolific where participants were only shown 1 critical sentence. Items were randomly from the 8 critical conditions in the stimuli pool, representing the 8 possible orders.

\paragraph{Materials.} We reduced the number of filler items and catch trials from 4 to 3, but increased their length and designed them to require more effort to further encourage attentive participants and clean data.
We further removed the bridge between S1 and S2 to be consistent with LM experiments.

We recruited a total of 242 participants, all of whom provided informed consent. We excluded 38 participants who failed at least one catch trial, which leaves 204 in the analysis.

\paragraph{Results.}
For more statistical power, we pooled responses from critical-first trials in Experiment 1a despite stimuli differences, yielding a total of $n = 326$. On the combined results, participants produced the repeated item in 60.1\% (196/326) of critical trials.

Breaking down by structure, we further examine ordering effects. Collapsing across other factors, the rate of producing X was numerically higher when the second disjunction order matched (ZX: 62.1\% vs. XZ:\ 58.2\%),
and when half order matched (h1h2: 61.9\% vs.\ h2h1: 58.4\%).
But accuracy was higher when the first disjunction order is mismatched (XY: 58.0\% vs.\ YX: 62.2\%).
None of these contrasts were significant.
We fit a mixed effect logistic regression predicting a binary indicator variable for ``correctly'' entering the continuation X, with one fixed effect term for whether the order fully matched in S1 and S2, and random effects by item with a random slope for the match term. We did not find a significant result ($\beta = -0.179$, $p = 0.647$). 
While it is possible that humans genuinely do not show ordering effects, we also think it plausible that Prolific participants were working quickly and that the data is noisy (as evidenced by the relatively low rate of generating the target noun even in the full match case).

\subsection{Discussion.} 
Experiment 1 establishes that the non-redundant disjunction phenomenon is robust in humans and LMs. The core critical vs.\ control contrast is significant. But LMs and humans diverge on ordering: LMs show strong sensitivity to the second disjunction's surface order matching between S1 and S2, while humans show null effects. 
While it is possible this is a genuine difference, there may be noise in the human data obscuring possible order effects.
The scaling pattern is also notable: the phenomenon seems to require models above roughly 400M parameters, suggesting that contextual binding required for non-redundancy is a capacity that emerges with scale.

\section{Experiment 2: Causal Intervention Shows Disjuncts Get Bound to Context}

Having established the behavioral pattern, we turn to exploring it mechanistically in LMs, in our effort to give a ``neural'' account of the phenomenon.
Our first hypothesis is that representations of key nouns take on relevant contextual information.
That is, for the sentence in \ref{ex:critical}, the ``France'' in ``France or Spain'' becomes contextually bound to ``a philosophy program''. And the ``France'' in ``Germany or France'' becomes contextually bound to ``a mathematics program''.
If LMs represent this contextual binding, it can provide a mechanism by which they can know the completion ``France'' is non-redundant.

To study this, we use ``activation patching'' \citep[][\textit{i.a.}]{meng2022locating, wang2023interpretability, hanna2023does, merullo2024circuit}. 
This is a causal interpretability method, in which a model's activations during a forward pass are systematically altered, with the resulting change in outputs analyzed to determine causal effect. In particular, we operate on two inputs, termed the base, $b$, and source, $s$. At a given site, we intervene by overwriting the LM's activations for an input $b$ with the activation $a_s$ from input $s$. By comparing the resulting output with the non-intervened pass of $b$, we can quantify the causal effect of the patched representation.

In particular, we perform the following activation patching. We take stimuli of the form of \ref{ex:source} and \ref{ex:base} to be our $s$ and $b$.

\ex. Maria will go to France or Spain for physics, or Germany or France for math. She will go to \textbf{France}$_1$ or Spain or Germany or \textbf{France}$_2$.\label{ex:source}

\ex. Maria will go to France or Spain for physics, or Germany or France for math. She will go to France or Spain or Germany or France. She will go to \textbf{France} \label{ex:base}

\begin{figure}[htbp]
\centering
\begin{subfigure}{\linewidth}
    \centering
    \includegraphics[width=.9\linewidth]{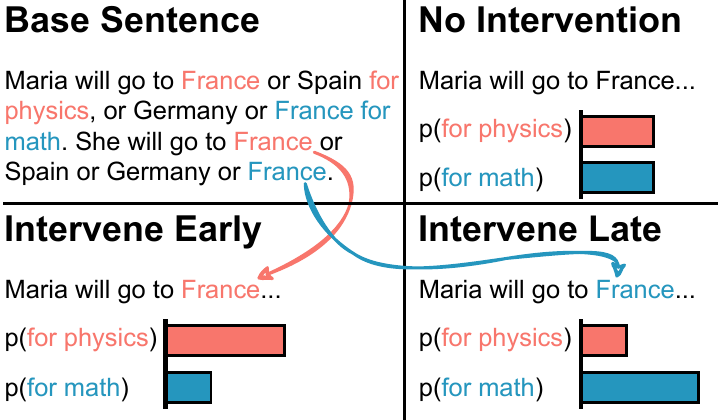}
    \caption{Schematic for our activation patching. We measure the probability of the contextually bound suffix when intervening with different instances of the repeated item.}
    \label{fig:patching-schematic}
\end{subfigure}

\vspace{0.25em}

\begin{subfigure}{\linewidth}
    \centering
    \includegraphics[width=.925\linewidth]{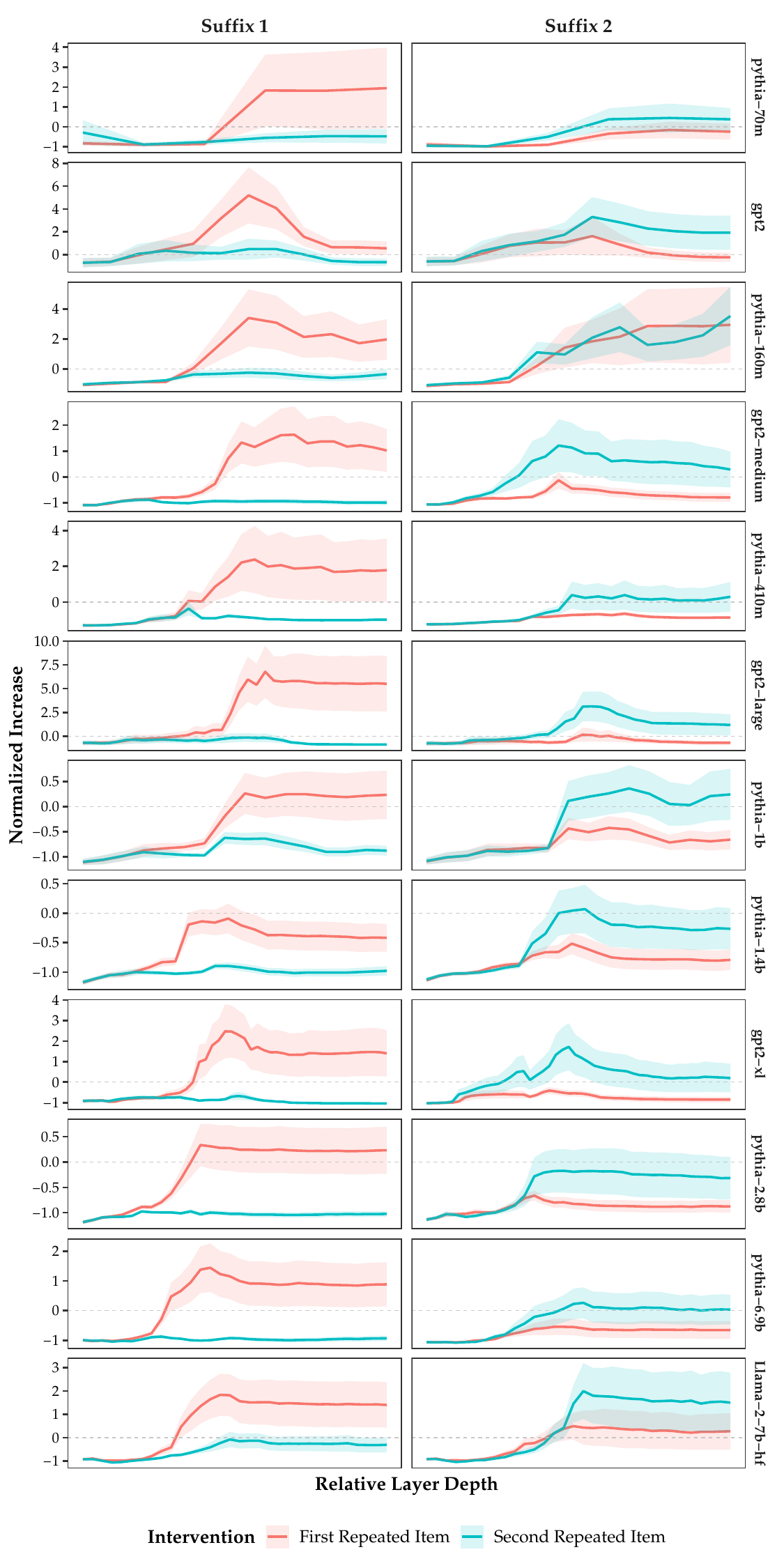}
    \caption{Patching across layers in each model.}
    \label{fig:patching-results}
\end{subfigure}
\caption{We perform activation patching across models and find disjunctive elements to be bound in different contexts.}
\label{fig:patching}
\end{figure}

In effect, we patch activations from each of the two `France' tokens in \textit{s} into the single ambiguous bolded `France' position in \textit{b}, then measure how the continuation probabilities (for ``physics'' vs. for ``math'') shift under intervention.
Were our LM to ``bind'' the disjunction to the context, we would expect to see that patching in the contextually different items from $s$ should disambiguate the repeated item in $b$, increasing the relative probability of the corresponding continuation under intervention.
Because the lexical identity of the patched token is held constant, differential effects across continuations must arise from contextual information encoded in the representation.
A schematic of this setup can be seen in Figure \ref{fig:patching-schematic}.

We sample critical items as described in the \nameref{sec:procedure} section, with the key distinction that S2 is of order YXZX, to ensure the first X we patch is unambiguous. We measure the model's token-level probabilities for each continuation directly and report the relative probability difference from the non-intervened baseline. This normalization allows easier comparisons across different continuations.

\paragraph{Results.} Figure \ref{fig:patching-results} shows our results. Across a wide range of models, we observe reliable continuation preferences under intervention. 
Specifically, patching in the repeated word from either of the disjunctions leads to a notable increase in the probability of the corresponding suffix in exactly the predicted direction. This suggests that the LMs have reliably ``bound'' the items within the disjunction to the surrounding context.

\paragraph{Discussion.}
This result provides causal evidence that LMs encode contextually distinct information in the representations of repeated tokens. The key finding is not merely that ``France'' is represented differently in different positions, but that the \textit{specific contextual information} (philosophy vs.\ mathematics) is encoded in a way that causally drives downstream predictions.

\section{Experiment 3: How Attention Heads Enable and Suppress Repetition}

We now turn to how the copying of the repeated item happens. 
To this end, we analyze how induction heads \citep{elhage2021mathematical} attend to the repeated item in both the critical and control settings. Induction heads are attention heads which concern themselves with the repetition of patterns. They have been found critical in various tasks including, but not limited to, in-context learning \citep{olsson2022incontextlearninginductionheads}, repeating items in lists \citep{merullo2024circuit} and outputting the indirect object in a sentence \citep{wang2023interpretability}. In this work, we analyze the most causally-relevant heads for induction in \texttt{Llama2-7b} \citep{touvron2023llama2openfoundation} as identified by \citet{feucht2025dualroute} using a sentence repetition task. 

We use the same stimuli as in Experiment 1 (no ``bridge''). For each of the identified induction heads, we compute the average attention at each of the S1 entity positions. 

\begin{figure}
    \centering
    \includegraphics[width=0.95\linewidth]{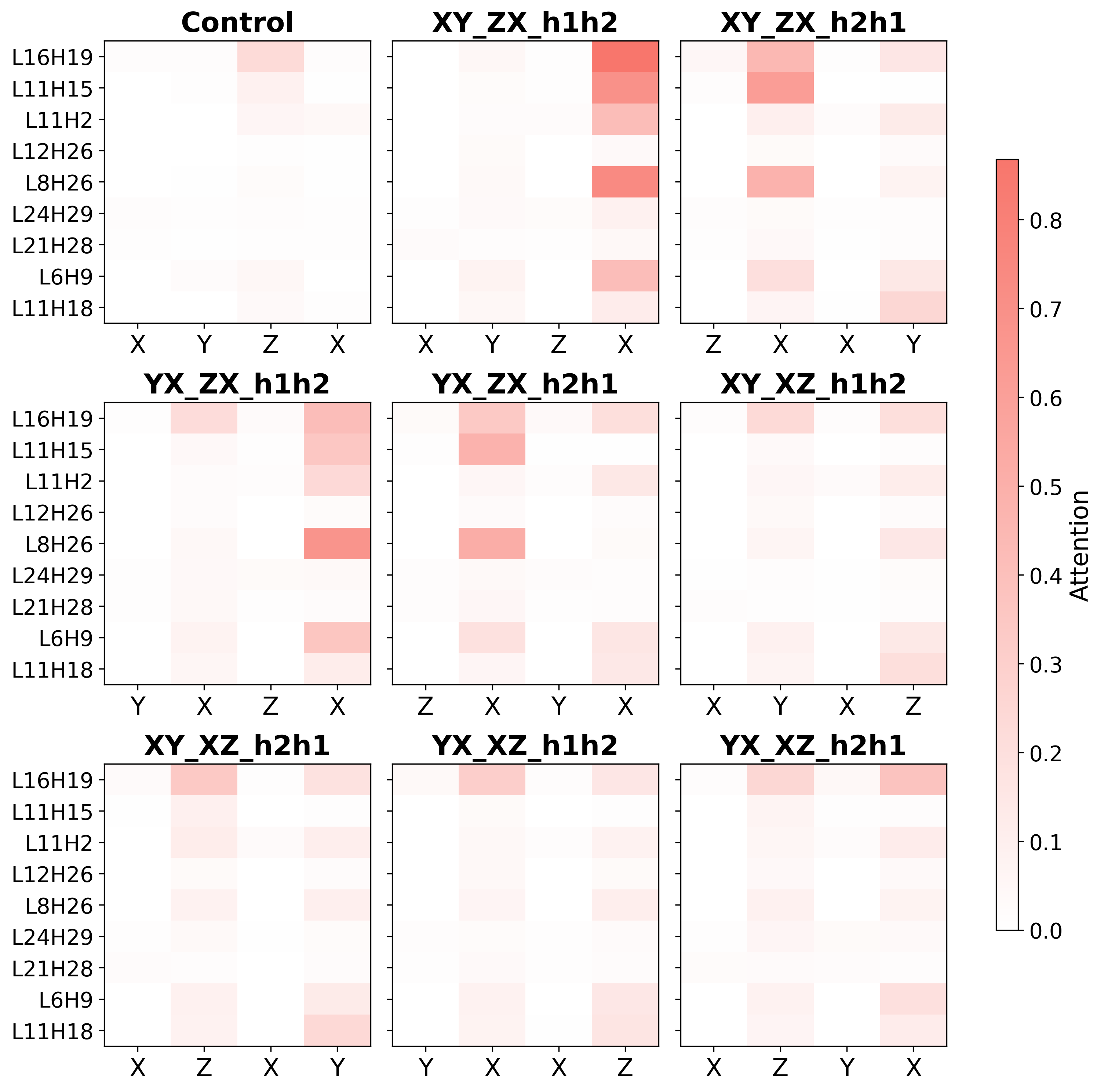}
    \caption{Average attention patterns of induction heads on the control condition and each of the critical conditions (with different entity orders). Rows = induction heads (top-9 by causal importance), columns = entity positions in S1, strength of color = average attention from the critical token to each position.}
    \label{fig:attention}
\end{figure}

\paragraph{Results.} 
We first compare attention patterns between the fully matching critical condition and the control. Our results are shown in the first two panels in Figure \ref{fig:attention}. 
In the critical case, we see a striking attention pattern at the key second X. In the control case, we see no such attention.
 
Further, we see induction heads are most strongly activated on the referent in the all-match critical condition (top row, middle), and other conditions that match in the second disjunction (ZX). 
Cases where the second disjunct are mismatched (XZ constructions; right middle row; bottom row) show much less activity.
This aligns with Figure \ref{fig:by-model-ordering}: models are most sensitive to entity ordering in the second disjunction.

\paragraph{Discussion.}
Experiment 2 showed \textit{what} information is encoded on each disjunct (contextual binding). Experiment 3 shows \textit{how} that information is deployed (selective attention by induction heads). The ordering sensitivity of the induction heads---strongest when the second disjunction's surface form matches between S1 and S2---provides a mechanistic explanation for the ordering effects in models.

\section{Discussion and Conclusion}
We have given empirical evidence for how both humans and LMs handle these peculiar disjunctions behaviorally. And we have given a neural analysis that explores, mechanistically, how LMs do it: activation patching shows that LMs causally bind repeated disjuncts to their surrounding context. Induction heads selectively attend to the contextually relevant instance to enable copying.

We see it as an open question how well this neural account fits in with existing formal semantic accounts. It is plausible that they might be isomorphic, but we also do not rule out potential differences; for instance, ordering effects complicate the picture. Mandelkern's analysis is order-invariant: the possibility conjuncts associated with ``$p$ or $q$'' are the same regardless of whether the surface form is ``$p$ or $q$'' or ``$q$ or $p$.'' LMs, by contrast, show strong sensitivity to the ordering of the second disjunction, in a way that is mechanistically grounded in induction heads that match surface patterns. 

So it remains an open question as to whether the LM representations encode something specifically analogous to Mandelkern's analysis.
Answering that question requires making more explicit the connection between our neural account and the formal  one. We leave this as a question for future work, but note that the nature of LM representations makes such an encoding in principle possible (and even natural). 

Zooming out, we believe theories afforded by a neural account valuably expand the linguist’s toolkit. In this specific case, this means grounding these disjunctions in attested and implemented mechanisms, predicting generalizations and systematic behaviors (like the ordering effects). More generally, however, the neural toolkit provides a system one can probe and perturb in order to generate theories and test them counterfactually. These all seem to us like desirable features of good linguistic explanations.

Is this kind of account explanatorily satisfying? 
We think so!
By being able to causally manipulate behaviorally competent LMs in a way consistent with our high-level symbolic theory, we aim to bridge the gap between distributional and formal semantic traditions, while mitigating weaknesses of each approach in isolation.

\section{Acknowledgments} KM acknowledges funding from NSF CAREER grant 2339729. QY acknowledges funding from the UT Harrington Fellowship. For helpful comments and discussions on these topics, we thank Josh Dever, Daniel Drucker, Matt Mandelkern, the UT Austin Computational Linguistics group, and the audience at the American Philosophical Association 2026 Central Division Meeting.

\section{AI Use Statement}

AI tools were utilized to pilot early LM behavioral results and further to aid in adapting our stimuli into the format required by Prolific for the human studies. Additionally, AI tools were used to aid in the adaptation of previous work's code for our Experiment 3. 
AI tools were also used for writing analysis code and summarizing results qualitatively.

\setlength{\bibhang}{0.125in}

\bibliographystyle{apalike}
\bibliography{actual}

\end{document}